# Bulyap Nesnelerini Tanıyıp Yerleştiren Raspberry Pi Tabanlı Akıllı Robot
# Raspberry Pi Based Intelligent Robot that Recognizes and Places Puzzle Objects

Yakup Kutlu[1], Zülfü Alanoglu[2], Ahmet Gökçen[1], Mustafa Yeniad[3]
[1]Bilgisayar Mühendisliği Bolumu, İskenderun Teknik Üniversitesi, Hatay, Türkiye
yakup.kutlu@iste.edu.tr, ahmet.gokcen@iste.edu.tr
[2]Bilgisayar Teknolojileri Bolumu, Mustafa Kemal Üniversitesi, Hatay, Türkiye
zalanoglu@mku.edu.tr
[3]Bilgisayar Mühendisliği Bolumu, Ankara Yıldırım Beyazıt Üniversitesi, Ankara, Türkiye
myeniad@ybu.edu.tr

*Özetçe—* **Bu çalışmada 3D modelleme programları ile tasarlanıp 3D yazıcıdan çıktı alınarak oluşturulan robot kol sistemi, Raspberry Pi ve USB kamera kullanılarak bilgisayar görme uygulaması gerçekleştirilmiştir. Sistem kameradan tespit edilen nesnelerin konum bilgilerini otomatik belirler ve konum bilgilerinden faydalanarak robot eklemlerinin yön ve açıları kontrol denklem çözüm yöntemi ile hesaplar. Yapılan hesaplamalar sonucunda robot kolunu hedefe yönlendirir. Kullanılan robot kolu 6 eksenli endüstriyel tiptir. Raspberry PI tabanlı geliştirilen sistemde python programlama dili ve görüntü işleme için OpenCV Kütüphanesi kullanılmıştır.**

*Anahtar Kelimeler—Raspberry PI; 3D tasarım; 6 eksen robot kolu; görüntü işleme; Python; OpenCV.*

*Abstract—* **In this study, computer vision application was realized by using Raspberry Pi, USB camera and 3D robotic arm system, which are designed with 3D modeling programs and output from 3D printer. The system automatically determines the position information of the detected objects from the camera and calculates the directions and angles of the robot joints using the position information. Because of calculations, the robot Arm moves to the target. The robot arm is 6-axis industrial type. The Raspberry PI-based system uses python programming language and OpenCV library for image processing.**

*Keywords— Raspberry PI; 3D modelling; 6 axis robot arm; image processing; Python; OpenCV.*

## I. GİRİŞ

Günümüzde özellikle tehlikeli işlerde, savunma sanayinde, insan sağlığını olumsuz etkileyecek ortamlarda robotların kullanılması önemli bir konu haline gelmiştir. Robotların insanlar gibi iş yapabilmesi insansı özellikler kazanmaları gerekmektedir. Öncelikle hedefin duyu organları ile tespit edilmesi, gerekli incelemeler, değerlendirmeler ve hesaplamalar yapılarak kolların harekete geçip eylemi gerçekleştirmesi gerekmektedir. Bu günlerde otonom robotlarda birçok özellik gelişmiştir. Örnek olarak; [1]'de belirtildiği gibi robotlar ile çevrenizi inşa edebilirsiniz ya da koruyabilirsiniz, [2] de ifade edildiği gibi planlama yaparak dinamik ortamlardaki yollarda kazasız yürütebilirsiniz ya da emniyetli bir şekilde insanlara nasıl yaklaşacağınızı planlayabilirsiniz[3]. Görmeye dayalı robot kolları görüntü ile kazanılmış verileri kullanır. Kameranın pozisyonuna ve manipülatörün özelliğine göre birtakım işlemler ile istenilen görevler yerine getirilir. En çok kullanılan alanların başında savunma ve askeri amaçlı sistemler, endüstriyel sistemler ve insan sağlığının olumsuz etkilenebileceği ortamlar gelir.

Güncel endüstriyel robot sistemleri yüksek çözünürlüklü görüntülere ve yapısal aydınlatmaya güvenerek nispeten basit görüntü işleme tekniklerini kullanır. Programlanabilen birçok araştırma yapılmasına rağmen yapay zeka destekli görmeye dayalı robot kolu uygulaması daha az yapılmıştır [4-6]. Robotun etkileşimde olduğu nesnelerin konumları değişken olabilir. Bu durumda nesnelerin yeri gerçek zamanlı değiştirildiğinde tüm sistem yeniden programlanmalıdır. [7] Her işlemde eklemlerin pozisyonu, hedefe ulaşmak için yeniden programlanır ve mümkün olduğunca hassas olur. Yapılan sistemin insan sistemine olabilecek en yakın seviyede olması robotik uygulamalarda temel amacıdır.

Özellikle 3D yazıcıların kullanılmaya başlamasından sonra robot tasarımları ve üretimleri kolaylaşmıştır. Sanal ortamda 3 boyutlu görsel çizim programları yardımıyla çizilen ve simüle edilen sistemler 3D yazıcılar yardımıyla fiziki olarak oluşturulup sabit görev tanımlı programlamayla her türlü alanda kullanılabilmektedir. Günümüz ve gelecek robot sistemlerinin akıllı ve kısmen özerk olmaları gerekmektedir. Yani yeni bir işi yapmaları için tekrar programlanmaları gerekmez [9]. Kameralardan ve diğer sensörlerden aldıkları verileri otonom olarak işler ve bir karara ulaşarak robot eklemlerini karar doğrultusunda hareket ettirirler.

Bu çalışmada görmeye dayalı akıllı robot kol kontrol sistemi sunulacaktır. 3D tasarlanıp oluşturulan robotik kol sistemi Raspberry PI ile kontrol edilir. Logitech C310 web kamerasından gerçek zamanlı alınan görüntü işlenip nesnelerin koordinatları belirlenerek robot kolun eklem açıları hesaplanıp ve hareket etmektedir. Burada önemli olan kısım robot kolunun önceden tanımlanmış görev fonksiyonları yoktur. Her işlem gerçek zamanlı olarak işlenip harekete geçmektedir.

## II. METARYAL VE METOTLAR

Kullanılan robot kolu 6 eksenli 3D tasarlanmıştır. Erimiş depozisyon modellemesinde (FDM) kullanılan iki yaygın filament türü Polilaktik Asit (PLA) ve Akrilonitril bütadien stirenidir (ABS) [12]. Çalışmada kullanılan robot dayanıklılık ve hafiflik temel alınarak ABS filamentden yapılmıştır. 3D tasarlanmış ve XYZ da Vinci 1.0 AIO marka Şekil 1'de gösterilen altı eksenli robot kolu tasarımı, 3D yazıcıdan çıktı alınmıştır. Raspberry PI geliştirme kartı, Logitech C310 web kamerası, ekran, klavye ve fare ile birlikte 5V volt 10 amperlik güç kaynağı kullanarak görmeye dayalı robot kolu tasarlanmıştır. Robotta 7 adet servo motor (Power HD 1501MG motorlar) mevcuttur. Bu motorlar 60 gr ağırlığında 6V ile çalışabilen 17 kg/cm torka sahip motorlardır. Tasarlanan sistem Şekil 2'de bir Raspberry Pi 2 görmeye dayalı robot kol uygulamasının blok şeması sunulmaktadır.

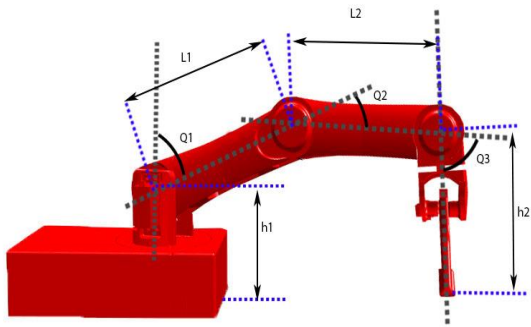

**Şekil 1.** Robot Kol Parçalarının Tasarım Grafiği.

Raspberry PI' nin tercih edilmesinin bir kaç sebebi vardır. Bunlardan bazıları: Çalışması için fazla güç gerekmez, çok düşük maliyetlidir (yaklaşık 35$). Lisans maliyeti gerektirmemekte, Debian tabanlı açık kaynak kodlu Rasbian işletim sistemi üzerinde çalışmaktadır. Üzerinde çeşitli amaçlar için kullanılabilecek 40 GPIO pin mevcuttur, Raspberry PI üzerindeki GPIO pinlerinden motora gönderilecek veriler alındığından motorlar için ayrıca bir motor sürücü devresi kullanılmamıştır.

### A. Robot Görme ve Açılarının Hesaplanması

Robot hareket sistemi iki aşamadan oluşmaktadır. Birinci aşama nesnenin görüntü işleme yöntemleri ile tespit edilip yerinin saptanması, ikinci aşama da robot açılarının hesaplanıp hareketin sağlanmasıdır.

Kamera verisinin işlenmesi, hedefin tespit edilmesi, koordinatlarının belirlenmesi ve robot kolunun belirtilen koordinata hareket etmesi için eklem açıların hesaplanması konusunda birçok araştırmacının ortaya koyduğu çalışmalar vardır [10-11]. Kameradan alınan görüntüde bulunan hedefin tespit edilmesi söz konusu olduğunda HAAR Cascade, Template Matching (Şablon Eşleştirme), (LBP) Local Binary Pattern, (HOG) Histogram of Oriented Gradients yöntemi gibi farklı yaklaşımlardan söz edilebilmektedir.

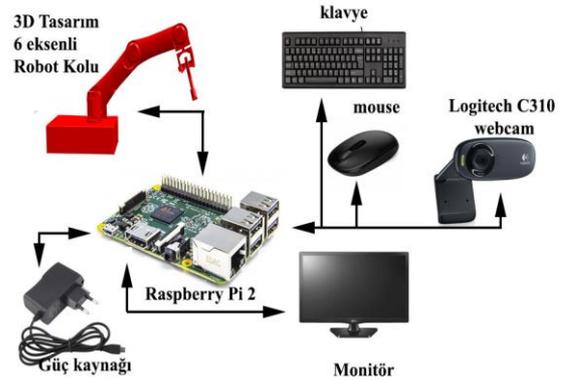

**Şekil 2.** Test kurulumunun blok şeması.

Birinci aşama olan nesnenin görüntü işleme yöntemleri ile tespit edilip yerinin saptanması için öncelikle üzerinde işlem yapılacak olan ilgi bölgesi (ROI) belirlenmiştir. Belirlenen ilgi bölgesi sırasıyla gri tona dönüştürülme, histogram eşitleme, siyah beyaz resme dönüştürme, kenar bulma ve resim iyileştirme işlemlerinde geçirilerek nesnelerin temiz bir şekilde belirlenmesi sağlanır [13]. Bu nesnelerin alanı belli bir değerden küçük olanlar silinerek resim üzerindeki gürültü olarak beliren objeler temizlenmektedir. Daha sonar her bir objenin ağırlık merkezleri belirlenerek kol hareketi için hedef belirlenmiş olur.

İkinci aşama da robot açılarının hesaplanıp hareketin sağlanması aşamasında öncelikle açıların belirlenmesi gerekmektedir. Robot kol açılarının belirlenmesinde geometrik çözüm yöntemlerinden Kontrol Denklem Çözüm Yöntemi kullanılmıştır. [14]

Robot kol açılarının hesaplamak için kontrol denklemlerin oluşturulması gerekmektedir. Ters kinematik yöntemlerinden kapalı form yaklaşım çözümleme yöntemi ile kontrol denkleminin hesaplaması yapılmıştır. Bu denklem için giriş parametreleri; L1, L2, L3 robot kol boyları, Çıkış parametreleri; Q1 Taban (Bel) Açısı, Q2 Omuz Açısı, Q3 Dirsek Açısı, Q4 Bilek Açısı, Q5 tutucu (Gripper) açısı şeklindedir olarak tanımlanmıştır.

Kontrol denkleminin hesaplanmasında robotun tutucu parmakların ulaşması gereken son noktaya göre robotun diğer eksenlerinin alacağı pozisyonların açısal değeri hesaplanmıştır. Bunun için kontrol denkleminin hesaplanmasında ters kinematik yöntemlerinden kapalı form yaklaşım çözümleme çeşidi kullanılmıştır. Sistemimiz gerçek zamanlı çalıştığı için kapalı form yaklaşımlardan biri olan geometrik çözüm metodu ile gerçekleştirilen hesaplama tutucu hariç 5 eksenli bir manipülatörün matematiksel hesaplaması olacaktır. Robotun 5 ekseni de döner eklemlerden oluşmaktadır.

Çalışmamızda robotun tutması gereken nesne yatay düzlemde bulunduğundan öncelikle taban (bel- Q1) açısını bulmamız gerekmektedir. Bunun için Şekil 3'deki düzlemin kuşbakışı görünümü ve uzunluk tanımlarından yararlanılır.

L1 ve L2'nin değerleri;

$$L1=(y1-y2) \ ve \ L2=(x2-x1) \quad (1)$$

şeklinde hesaplanır.

L3 değeri ise hipotenüs teoremi yardımıyla;

$$L3=\sqrt{l1^2 + l2^2} \quad (2)$$

olarak bulunur.

Kenar uzunlukları bilinen bir üçgende istediğimiz açıyı $cos^{-1}$ yardımıyla denklem (3) deki gibi bulunabilir.

$$Q1= cos^{-1}((l1^2+l3^2-l2^2)/(2*l1*l3)) \quad (3)$$

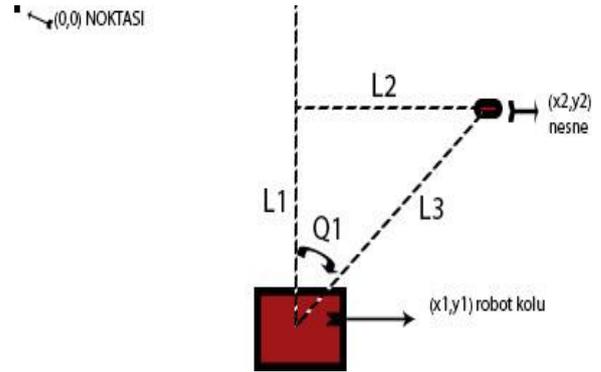

**Şekil 3.** Düzlemin kuşbakışı görünümü ve uzunluk tanımları

Buradaki L3 uzunluğu bulmak için öncelikle kameradan alınan görüntü Python programlama dilinde OpenCv kütüphanesini kullanarak nesne ile robotun koordinatları bulunur daha sonra bu koordinatlar arasındaki uzaklık hesaplanır.

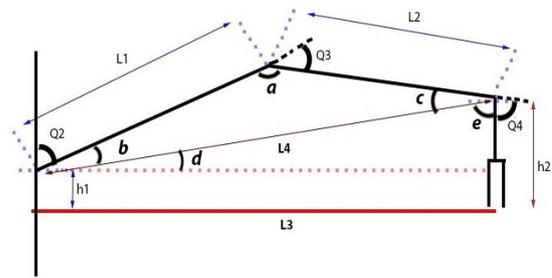

**Şekil 4.** Robot kol uzunluk ve açıları

Şekil 4'de de görüldüğü gibi robot kol uzunlukları olan L1 ve L2 uzunlukları, robot tabanı ile nesne arasındaki uzunluk olan L3 uzunluğu, robot tabanının zeminden yüksekliği olan h1 ve robot bileğinin zeminden yüksekliği olan h2 bilinmektedir. Burada L4 uzunluğu hipotenüs teoremi ile;

$$L4=\sqrt{(h2-h1)^2 + l3^2} \quad (4)$$

olarak bulunur.

Şekildeki tüm uzunluklar bilindiğine göre a, b, c, d, e açılarını bulabiliriz.

$$a= cos^{-1}((l1^2+l2^2-l3^2)/(2*l1*l2)) \quad (5)$$

$$b= cos^{-1}((l1^2+l4^2-l2^2)/(2*l1*l4)) \quad (6)$$

$$c= cos^{-1}((l2^2+l4^2-l1^2)/(2*l2*l4)) \quad (7)$$

$$d= cos^{-1}((l4^2+l3^2-(h2-h1)^2)/(2*l4*l3)) \quad (8)$$

$$e= cos^{-1}(((h2-h1)^2+l4^2-l3^2)/(2*(h2-h1)*l4)) \quad (9)$$

şeklinde hesaplanır.

a, b, c, d, e açıları hesaplandıktan sonra kollarda bulunan motorların asıl dönmesi gereken açılar olan Q2, Q3, Q4 açıları;

$$Q2=90-(b+d) \qquad (10)$$

$$Q3=90-a \qquad (11)$$

$$Q4= 180-(c+e) \qquad (12)$$

olarak bulunur.

## III. DENEYSEL UYGULAMA

Çalışmada Şekil 5 da görüldüğü gibi deneysel düzenek hazırlanmıştır. Robot kolu, kontrol birim (Raspberry Pi) ve nesneleri yukarıdan görmek için bir adet sabit USB kamera mevcuttur. Nesnelerin yerini tespiti sırasında kamera görüntüsünün engellenmemesi için robot kolu dik pozisyona getirilmiştir.

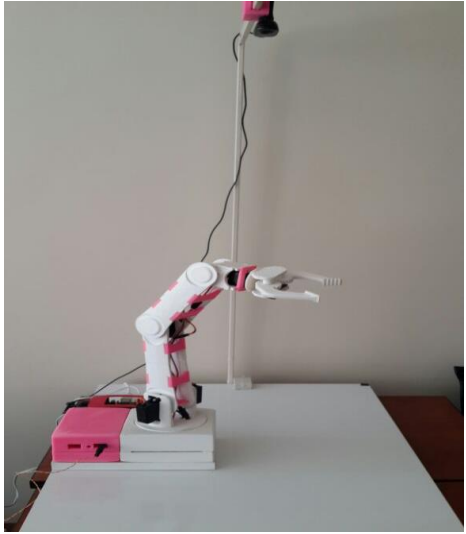

Şekil 5. Deneysel Düzenek

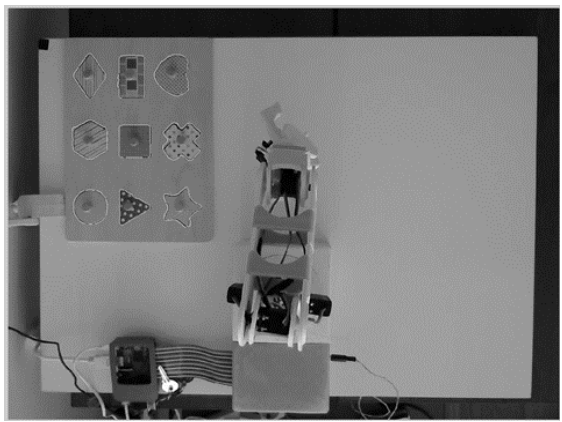

Şekil 6. Bul-tak nesneleri kamera Görüntüsü

Deneysel uygulamada kullanılacak bul-tak nesnelerinin hazır durumdaki kamera görüntüsü Şekil 6'daki gibidir. Kameradan ekran görüntüsü alındıktan sonra görüntü işleme başlatılarak. Şekil 7 'de görüldüğü gibi sırasıyla; Resimde görüntü işlemede kullanılacak ilgi bölgesi (ROI) belirleme, gri resme dönüştürme, histogram eşitleme, kenar bulma ve filtreleme işlemleri yapılmıştır. Son oluşan resimdeki belirlenen objelerin her birinin ağılık merkezi bulunmuş ve daire içerisine alınarak işaretlenmiştir.

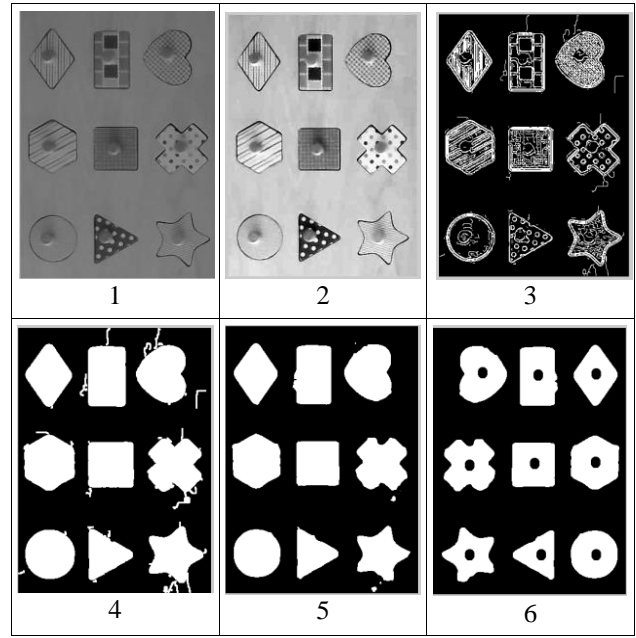

Şekil 7. Görüntü işleme süreçleri ve sonuçları

Bu deneysel çalışmada kullanılan yapboz nesnelerini belirlenen ağırlık merkezlerinin koordinatları kullanılarak robot kol eklem açıları hesaplanmıştır. Robot kol, her bir bul-tak nesnesini alarak görev gereği diğer tarafa karışık halde yerleştirmesi sağlanmıştır. Şekil 8 de robot kol görevi tamamladıktan sonar elde edilen görüntü verilmiştir.

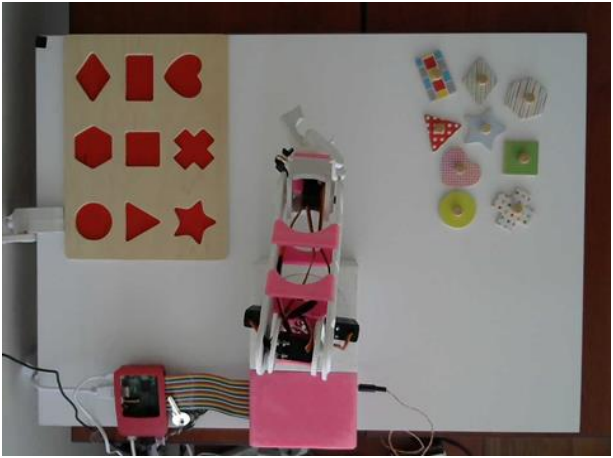

Şekil 8. Bul-tak nesneleri görüntü işleme sonucu robot kol ile yer değiştirme sonucu

## IV. SONUÇ VE TARTIŞMA

3D yazıcı ile oluşturulan 6 eksenli robot kolu ve web kamerası kullanılarak görmeye dayalı robot kolu kontrol sistemi geliştirilmiştir. Raspberry Pi temelli sistemde, tüm işlemler gömülü sistem üzerindeki işletim sistemi ve geliştirilen yazılımlar ile yapılmıştır.

Çalışmada kullanılan Debian tabanlı Rasbian işletim sisteminin de birçok avantajı mevcuttur. Açık kaynak olduğu için lisans gerektirmez ve geliştirilebilir bir sistemdir. Yazılım geliştirme tarafında Python dili kullanılması ve python scriptlerinin derlenmesine gerek duyulmadan çalıştırılabilmesi rahatlıkla başka sistemlerde kullanılabilir bir alt yapı oluşturmaktadır. Ayrıca OpenCV Kütüphanesi gömülü sisteme bütünleşmiş edilerek kabiliyeti arttırılmıştır.

İşlenen görüntü sonucu tespit edilen nesnelerin taşınması için robot kolun eklem hareketlerinde kullanılan ters kinematik yöntemlerinden biri olan geometrik çözüm metodu sisteme başarı ile uygulandığı görülmüştür. Robot tutucuların, nesnelerin üzerindeki ağırlık merkezlerinde bulunan noktaları tutması ile taşıma gerçekleştirilir. Fakat nesneleri tutup tutmadığını algılayacak bir denetim mevcut olmadığından nesnelerin düşmesi söz konusu olmuştur. Bu sebepten dolayı tutucuların sensörler ile desteklenmesi gerektiği ve gelecek çalışmalarda yapılması düşünülmektedir.

## KAYNAKÇA

[1] H. Durrant-Whyte, T. Bailey, "Simultaneous localization and mapping: part i, Robotics Automation Magazine", IEEE 13 (2) (2006) 99 –110. doi:10.1109/MRA.2006.1638022.

[2] S. Petti, T. Fraichard, "Safe motion planning in dynamic environments, in: Intelligent Robots and Systems", (IROS 2005). 2005 IEEE/RSJ International Conference on, 2005, pp. 2210 – 2215. doi:10.1109/IROS.2005.1545549, 2005.

[3] S. Satake, T. Kanda, D. Glas, M. Imai, H. Ishiguro, N. Hagita, "How to approach humans?-strategies for social robots to initiate interaction, in: Human-Robot Interaction (HRI)", 4th ACM/IEEE International Conference on, 2009, pp. 109 –116, 2009

[4] D. Harwood A. Elgammal and L. Davis, "Non-parametric model for background subtraction" In Proceedings of the European Conference on Computer Vision, pages 751-767, 2000.

[5] R. Brunelli. "Template Matching Techniques in Computer Vision: Theory and Practice". John Wiley and Sons, 2009.

[6] L. Xingzhi and S. M. Bhandarkar." Multiple object tracking using elastic matching" In Proceedings of the IEEE Conference on Advanced Video and Signal Based Surveillance, pages 123-128, 2005.

[7] H. Guo-Shing, C. Xi-Sheng, C. Chung-Liang, "Development of dual robotic arm system based on binocular vision", International Automatic Control Conference, pp. 97-102, 2013.

[8] M. Seelinger, E. Gonzalez-Galvan, M. Robinson, S. Skaar, "Towards a robotic plasma spraying operation using vision", *IEEE Robotics & Automation Magazine*, vol. 5, no. 4, pp. 33-38, 1998.

[9] V. Lippiello, F. Ruggiero, B. Siciliano and L. Villani "Visual Grasp Planning for Unknown Objects Using a Multifingered Robotic Hand" IEEE/ASME Transactions on Mechatronics, vol. 18, no. 3, pp. 1050-1059, 2013

[10] B.Iscimen, H.Atasoy, Y.Kutlu, S.Yildirim, E.Yildirim "Bilgisayar Görmesi ve Gradyan İniş Algoritması Kullanılarak Robot Kol Uygulaması" Akıllı Sistemlerde Yenilikler ve Uygulamaları Sempozyumu (ASYU 2014) (Yayın No:1558815,2014

[11] B.Iscimen, H.Atasoy, Y.Kutlu, S.Yildirim, E.Yildirim "Smart robot arm motion using computer vision" Elektronika ir Elektrotechnika 21.6 (2015): 3-7,2015

[12] F. Decuir ,K.Phelan, B. Hollins "Mechanical Strength of 3-D Printed Filaments" Biomedical Engineering Conference (SBEC), 2016 32nd Southern, March 11-13, Shreveport, LA, USA,2016

[13] http://docs.opencv.org/3.0-beta/doc/py_tutorials/py_imgproc/py_morphological_ops/py_morphological_ops.html

[14] M.Chen, Y.Gao "Closed-Form Inverse Kinematics Solver for Reconfigurable Robots" International Conference on Robotics & Automation Seoul, Korea May 21-26, 2001